# GNC OF THE SPHEREX ROBOT FOR EXTREME ENVIRONMENT EXPLORATION ON MARS

Himangshu Kalita[*], Ravi Teja Nallapu[†], Andrew Warren[‡], and Jekan Thangavelautham[§]

Wheeled ground robots are limited from exploring extreme environments such as caves, lava tubes and skylights. Small robots that can utilize unconventional mobility through hopping, flying or rolling can overcome these limitations. Multiple robots operating as a team offer significant benefits over a single large robot, as they are not prone to single-point failure, enable distributed command and control and enable execution of tasks in parallel. These robots can complement large rovers and landers, helping to explore inaccessible sites, obtaining samples and for planning future exploration missions. Our robots, the SphereX, are 3-kg in mass, spherical and contain computers equivalent to current smartphones. They contain an array of guidance, navigation and control sensors and electronics. SphereX contains room for a 1-kg science payload, including for sample return. Our work in this field has recognized the need for miniaturized chemical mobility systems that provide power and propulsion. Our research explored the use of miniature rockets, including solid rockets, bi-propellants including RP1/hydrogen-peroxide and polyurethane/ammonium-perchlorate. These propulsion options provide maximum flight times of 10 minutes on Mars. In addition, we have been developing mechanical hopping mechanisms. Flying, especially hovering consumes significant fuel; hence, we have been developing our robots to perform ballistic hops that enable the robots to travel efficiently over long distances. Techniques are being developed to enable mid-course correction during a ballistic hop. Using multiple cameras, it is possible to reconstitute an image scene from motion blur. Hence our approach is to enable photo mapping as the robots travel on a ballistic hop. The same images would also be used for navigation and path planning. Using our proposed design approach, we are developing low-cost methods for surface exploration of planetary bodies using a network of small robots.

---

[*] PhD Student, Space and Terrestrial Robotic Exploration Laboratory, Arizona State University, 781 E. Terrace Mall, Tempe, AZ.
[†] PhD Student, Space and Terrestrial Robotic Exploration Laboratory, Arizona State University, 781 E. Terrace Mall, Tempe, AZ.
[‡‡] Undergraduate Student, Space and Terrestrial Robotic Exploration Laboratory, Arizona State University, 781 E. Terrace Mall, Tempe, AZ.
[§§] Assistant Professor, Space and Terrestrial Robotic Exploration Laboratory, Arizona State University, 781 E. Terrace Mall, Tempe, AZ.

**INTRODUCTION**

Mobile ground robots have become integral for surface exploration of Moon, Mars and other planetary bodies[3]. Past missions have inferred the presence of water on the Moon[1] and evidence of past water flow on Mars[2]. These rovers have proven their merit but are large, in the order of 200 kg to 800 kg or more and are sophisticated, housing state of the art science laboratories. With rapid advancement in lightweight structural materials, miniaturization of electronics, sensors and actuators it is possible to develop small, lightweight and low-cost platforms to tackle some of the hardest challenges in planetary surface exploration.

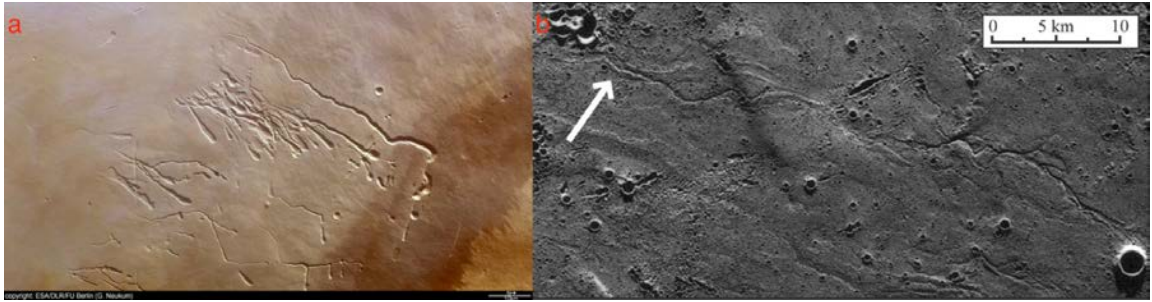

**Figure 1. Lava tubes on the surface of Mars and Moon. (a) Lava tubes observed in Pavonis Mons on Mars. (b) Lava channel observed in the southern Imbrium basin on Moon[5].**

Mobility is a critical element in solving the upcoming challenges of planetary surface exploration. Although wheeled ground robots have excellent performance on relatively flat, benign, even terrains, their obstacle traversing capabilities are typically limited to wheel diameter and they are limited from exploring extremely rugged environments such as caves, lava tubes and skylights. Developing small, cost-effective robots that can utilize unconventional mobility through hopping, somersaulting, flying or rolling can overcome these limitations. When these mobility solutions are scaled up to groups of robots or swarms, a large area maybe covered in short duration[3]. Multiple robots, operating as a team offer significant benefits over a single large rover, as they are not prone to single-point failure, enable distributed command and control and enable exploration in parallel. These robots can complement large rovers and landers, helping to explore inaccessible sites, obtaining samples and for planning future exploration missions.

In this paper, we present a new spherical robot called SphereX. SphereX has a mass of 3 kg and contains electronics and sensors equivalent to current smartphones. Each robot also contains an array of guidance, navigation and control sensors and volume for a 1 kg science payload. We consider use of combined power and propulsion systems using chemical energy. In this paper, we focus on use of chemical rockets for propulsion. We provide a comparative study between solid propellants, liquid bi-propellants and liquid monopropellants based on $I_{sp}$ and flight time. Flying, especially hovering consumes significant fuel; hence we have sought alternative solutions that improve on fuel use and range. Ballistic hops overcome obstacles that maybe many times larger than the robot, enabling short flights, while also providing range. Ballistic hops are not as efficient as rolling. However, they enable traversing rugged terrain where conventional wheel robots may get stuck. In this work, we model the dynamics of these hopping robots and propose guidance, navigation and control solutions.

SphereX requires use of a propulsion system and Attitude Determination and Control System (ADCS) to perform controlled ballistic hops. The propulsion system consists of 4 thrusters that generate thrust along the robot's +$z$ axis. Differential throttling of the four thrusters generates a torque that maybe used by the system to change its roll, pitch and yaw angles. The 3-axis miniature reaction wheels enable the robot to spin/pan to take images. Navigation is done using the onboard stereo cameras that detect obstacles, measure its distance and autonomously generate trajectories to avoid them. Once a SphereX robot is on the ground, the onboard cameras would take precise, high-resolution panoramic and stereo panoramic images. These images may be used to map the surrounding for navigation, path planning, mission science and public outreach. Stereo video generated during flight can be post-processed to be viewable using VR (Virtual Reality) head-sets, giving the public and mission planners the impression of being on site.

To explore a cave or lava tube requires a team of SphereX robots that work collaboratively to map, navigate and communicate the data back to the base station. Often, there will not be line of sight communication between the base station and the robot team. Hence, the robots need to act as relays to pass messages from the base station to individual robots along the cave much like a bucket-brigade. A comparative feasibility study of the time taken to map various cave with respect to distance travelled and image resolution is illustrated. Our studies show that it is possible to develop low-cost methods to surface and subsurface exploration of planetary bodies using a team of small robots. In the following sections, we present background and related work, followed by a system overview of the SphereX robot, presentation on the propulsion and the dynamics of robot hopping, followed by discussions, conclusions and future-work.

**BACKGROUND AND RELATED WORK**

In this section we review various proposed techniques for extreme-environment mobility and exploration. Small spherical robots have been widely proposed. Their spherical shape enables them to roll on loose, even terrain. Examples include spherical robots developed at Univ. of Sherbrooke[7], Kickbot[8] developed at MIT, Cyclops[9] at Carnegie Mellon University and inflatable ball robots developed at North Carolina State University[25] and University of Toronto[26]. Typically, these spherical robots use a pair of direct drive motors in a holonomic configuration[33]. Others such as the Cyclops[9] and the inflatables pivot a heavy mass, thus moving center of gravity that results in rolling. Other mobility techniques including use of spinning flywheels attached to a two-link manipulator on the Gyrover[10] or 3-axis reaction wheels to spin and summersault as with the Hedgehog developed by Stanford and NASA JPL[27]. Hedgehog's use of reaction wheels enables it to overcome rugged terrain by simply creeping over the obstacle no matter how steep or uneven[27]. However, it's unclear if a gyro based system can overcome both steep and large obstacles. In reality, even a gyro based system is bound to slip on steep surfaces, but under low gravity environments such as asteroids, they may be able reach meters in height.

An alternative to rolling and creeping is hopping. A typical approach to hopping is to use a hopping spring mechanism to overcome large obstacles[28]. One is the Micro-hopper for Mars exploration developed by the Canadian Space Agency[6]. The Micro-hopper has a regular tetrahedron geometry that enables it to land in any orientation at the end of a jump. The hopping mechanism is based on a novel cylindrical scissor mechanism enabled by a Shape Memory Alloy (SMA) actuator. However, the design allows only one jump per day on Mars. Another technique for hopping developed by Plante and Dubowsky at MIT utilize Polymer Actuator Membranes (PAM) to load a spring. The system is only 18 grams and can enable hopping of Microbots with a mass of 100 g up to a 1 m[11, 12]. Microbots are cm-scale spherical mobile robots equipped with power and communication systems, a mobility system that enables it to hop, roll and bounce and an array of miniaturized sensors such as imagers, spectrometers, and chemical analysis sensors developed at MIT[11, 12]. They are intended to explore caves, lava-tubes, canyons and cliffs. Ideally, many hundreds of these robots would be deployed enabling large-scale in-situ exploration. Mapping and localization of cave environments using familiar techniques such as Simultaneous Localization and Mapping (SLAM) have been shown recently[13, 14]. However, current techniques still don't account for the limited lighting conditions.

SphereX is the direct descendant of the Microbot platform. SphereX has the same goals as the Microbots, but with the goal of launching fewer robots, that are better equipped with science-grade instruments. ASU's SpaceTREx laboratory is also experimenting with mechanical hopping mechanisms for the SphereX platform and that uses a motor to tighten a torsional spring[29]. Other techniques for hopping mimic the grasshopper and use planetary gears within the hopping mechanism[30].

Mechanical hopping systems can use onboard electrical power, using batteries or PEM fuel cells. PEM fuel cells are especially compelling as techniques have been developed to achieve high specific energy, solid-state fuel storage systems that promise 2,000 Wh/kg[15,20,33]. Feasibility studies show that the robots can travel 10s of kilometers without recharging. However, for short, focused, simple missions, both the mechanical actuators and systems to power them all impose increased complexity and cost.

An interesting alternative to hopping is flying. In theory, flight provides a unique point of view above a terrain of interest and minimizes concerns of bypassing large or impassable obstacles. A few mission concepts have proposed methods to fly off-world. This includes NASA JPL's helicopter for Mars[31] and NASA Goddard's ARES rocket-powered aircraft[32]. Both systems have large footprints. Rocket are the most compelling option for flight in off-world environments with thin or no atmosphere. Our work has focused on miniaturizing rocket thrusters to enable flight using a small footprint.

**SYSTEM OVERVIEW**

In this section we present the SphereX spherical microbot that are capable of hopping, flying and rolling through caves, lava-tubes and skylights. Figure 3 shows the internal and external views of SphereX. The lower half of the sphere contains the propulsion system, with storage tanks for fuel and oxidizer connected to four main thrusters. The attitude control system is at the center of mass of the robot and contains a system of 3-axis reaction wheel for maintaining roll, pitch and yaw. The main thruster enables translation along +z axis and along with the attitude control system it enables the robot to move along 3 axes. Next is the Lithium Thionyl Chloride batteries arranged in circle as shown. A pair of stereo cameras and a laser range finder rolls on a turret. This enables the robot to take panoramic pictures and can scan the environment without having to move using the propulsion system. Moreover, the stereo camera and laser range finder would perform accurate navigation and perception. Above the turret are two computer boards, IMU and IO-expansion boards, in addition to a power board. The volume above the electronics is reserved for a science payload of up to 1 kg[4].

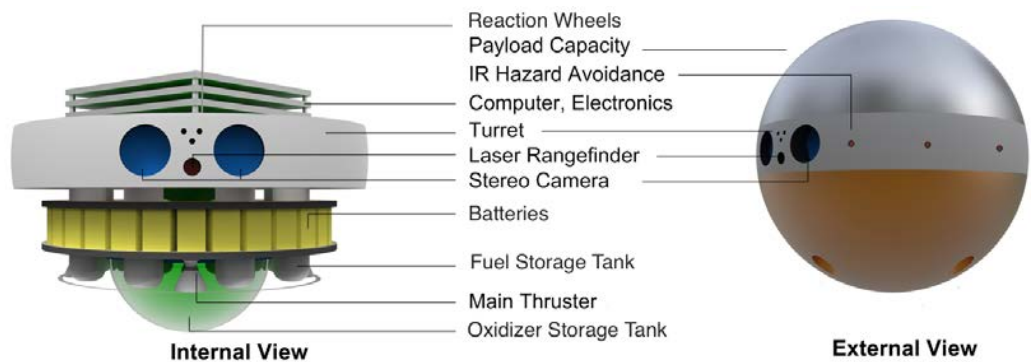

**Figure 2. Internal and External views of SphereX**

All of the hardware components, except the propulsion system can be readily assembled using Commercial Off-The-Shelf (COTS) components. We have analyzed several chemical propulsion systems, including solid propellants, liquid bi-propellants and liquid monopropellants. The proposed propulsion system uses RP-1 as fuel and $H_2O_2$ as the oxidizer. The mass budget for a single spherical robot is shown in Table 1. The sensors and electronics are relatively compact, with the bulk of the mass being occupied by propulsion and science payload.

**Table 1. SphereX Mass Budget**

| Major Subsystem | Mass (Kg) |
| --- | --- |
| Electronics | 0.2 |
| Power | 0.3 |
| Stereo Camera, Laser Rangefinder | 0.3 |
| Propulsion | 0.8 |
| ADCS | 0.4 |
| Payload | 1 |
| Total | 3 |

## SYSTEM MOBILITY

In this section, we analyze the SphereX mobility system. The mobility system uses miniature rockets. We consider solid rockets, bi-propellants including RP1/hydrogen-peroxide and polyurethane/ammonia-perchlorate. The primary attitude control system consisting of 3-axis reaction wheels aids in pointing the rocket thrusters in the right direction. Using this mobility system, the robot can perform multiple hops.

### Solid Rocket Motor

Solid rockets can be arranged until multiple pellets, with each pellet ignited once. Solid rockets work burning a solid grain that contain the fuel and oxidize in a solid grain, result in the formation of hot gasses that produce thrust. Figures 4 and 5 shows the profile for Isp, chamber pressure and thrust for 20% Polyurethane and 80% ammonium perchlorate by weight with ambient pressure 600 Pa, cylindrical solid rocket motor of inner radius 8 mm, outer radius 18 mm, length 15 mm and a nozzle of throat radius 2 mm.

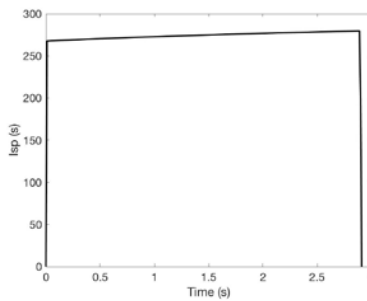

**Figure 4. Solid Rocket pellet $I_{sp}$ varying with time.**

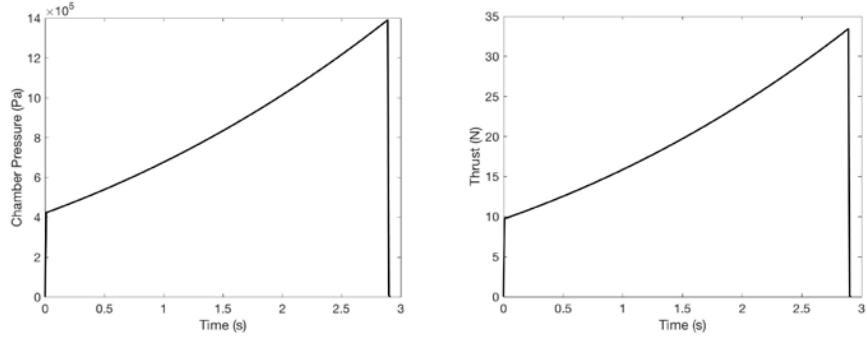

**Figure 5. Solid rocket pellet chamber pressure (left) and thrust evolution (right).**

**Liquid-Propellant Rocket Motor**

In addition to solid propellants, we consider liquid monopropellant that typically self-ignite or ignite upon contact with a catalyst. However, monopropellants such as hydrazine pose safety and handling concerns, while hydrogen peroxide requires high concentrations. Both options pose concerns when used in small robots that may experience shock and impact forces. Bi-propellants require a fuel and oxidizer source and are preferred for their high performance, throttle-ability and for safety in storage and handling. The challenge is finding bi-propellant combinations that offer high Isp without requiring cryogenic storage.

**Ballistic hop of the SphereX robot using Solid/Liquid-Propellant Rocket Motor**

The governing equations for the motion of the SphereX robot are expressed by the translational motion of its center of mass and the rotational motion of the body about its center of rotation. These equations include the translational kinetic, angular kinematic and angular kinetic equations. The angular kinetic equations consist of governing equations of the total system that include equations of the SphereX robot with the reaction wheels and equations of the reaction wheels alone[18]. Reaction wheels are momentum exchanging devices which operates on the principle of conservation of momentum, which states that the total momentum of a closed system is constant. Thus, the governing equations of the total system indicates that the attitude control system is highly nonlinear, and has three inputs which are the torques exerted by the reaction wheels and three outputs which are the desired Euler angles. So, to maintain the SphereX robot in its desired orientation we have designed a PD control algorithm that generate control torque inputs as a function of attitude errors as shown below:

$$\tau_{rw} = -K_p(e_{des} - e_{act}) - K_d(\omega_{des} - \omega_{act}) \qquad (4)$$

where $K_p$ and $K_d$ are the proportional and derivative control gains, $e_{des}$ and $e_{act}$ are the desired and actual Euler angles, $\omega_{des}$ and $\omega_{act}$ are the desired and actual angular velocity of the SphereX robot respectively. Figure 7, 8 and 9 shows the trajectory, Euler angles, and angular velocity of the SphereX robot for the PD control algorithm. The desired Euler angles were 0.27, 0.25 and 0.07 radians and the desired angular velocities were 0 rad/s. It is clear that the PD law is able to attain the desired Euler angles and angular velocities as commanded.

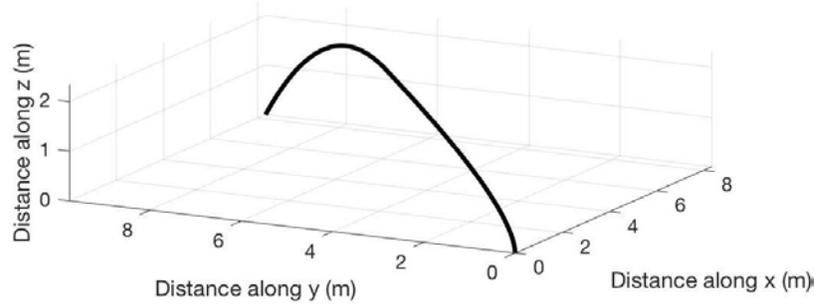

**Figure 6. Trajectory of the SphereX robot.**

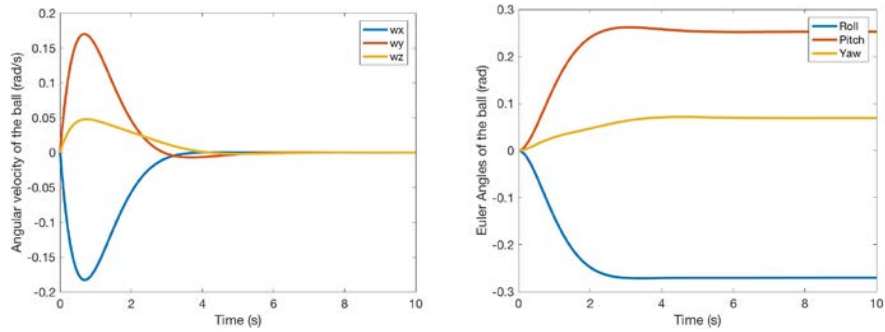

**Figure 7. (Left) Euler angles of the robot. (Right) Angular velocity of the robot.**

The flight time and mass of propellant required for a single hop of the SphereX robot is shown for various solid propellants. It is clear that CDT(80) propellant have the highest Isp of 325s and hence the highest flight time of 448 seconds and is competitive with bi-propellants.

**Table 2. Comparison of solid propellants**

| Propellant | Molecular Weight (Kg/kmol) | Combustion Temperature (K) | $I_{sp}$ (s) | Flight time (s) |
|---|---|---|---|---|
| JPL 540A | 25 | 2,600 | 280 | 360 |
| ANP-2639AF | 24.7 | 2,703 | 295 | 370 |
| CDT(80) | 30.18 | 4,000 | 325 | 448 |
| TRX-H609 | 25.97 | 3,040 | 300 | 398 |

Table 3 shows the comparison of various liquid monopropellants and bipropellants with respect to its specific impulse and flight time. Kerosene and hydrogen peroxide offer excellent performance, without the safety and handling concerns of hydrazine. Furthermore, the hydrogen peroxide maybe diluted below 50 % to minimize concerns of self-ignition.

**Table 3. Comparison of liquid propellants**

| Fuel | Oxidizer | Molecular Weight (Kg/kmol) | Combustion Temperature (K) | Isp (s) | Flight time (s) |
|---|---|---|---|---|---|
| Kerosene | $H_2O_2$ | 22.2 | 3,008 | 333 | 400 |
| Hydrazine | $HNO_3$ | 20 | 2,967 | 349 | 440 |

| (CH$_3$)$_2$NNH$_2$ | HNO$_3$ | 23.7 | 3,222 | 334 | 400 |
| H$_2$O$_2$ | | 22.7 | 1,278 | 214 | 260 |
| Hydrazine | | 10.29 | 966 | 277 | 340 |
| CH$_3$NO$_2$ | | 20.3 | 2,646 | 326 | 380 |

**Obstacle avoidance**

For sensing the obstacles, the robot must be able to calculate the obstacle-to robot distance and the dimensions (width and height) of the obstacles[19]. We will use machine vision techniques to sense obstacle distance and its dimensions. The SphereX robot is equipped with two CMOS cameras which takes images of the surrounding by the pinhole lens model and every point in a 3D space denoted by *M* is transformed into a pixel *m*. The relationship between 3D point *M* and its projected 2D point *m* is shown below [21]:

$$sm' = A[RT]M'  \qquad (5)$$

where *s* is a scaling factor, *R* is a 3×3 rotation matrix, *T* is a 3×1 translation vector and A is a 3×3 matrix that describe the internal characteristics of the camera. Calculating the real coordinates of each point of the obstacle, we can calculate the obstacle-to-robot distance and its dimensions. Based on the position of the gap and the states (position, velocity, and orientation) of the robot, the desired trajectory of the robot is determined. Photos taken by the SphereX robots are also used for navigation and path planning using potential fiends[22]. Once a navigation potential function is known the robot velocity is commanded along the negative gradient of the potential function. Figure 8 shows the contour plot of the navigation function with four obstacles centered at (-0.2, -0.35), (0.4, 0.46), (0.43, -0.38), (-0.58, 0.44) and radius 0.17, 0.19, 0.22, 0.24. The robot trajectory is shown in black color whose starting point is (0.6, 0.73) shown by the black dot and its goal is (-0.4, 0). This approach permits smooth movement to get from starting point to goal location while avoiding obstacles.

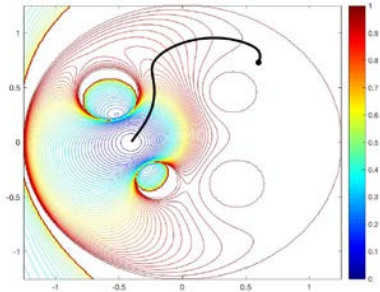

**Figure 8. Contour plot of the Navigation function and robot trajectory**

**Communication**

In cave environments, there is no line of sight from a starting point to some corridor or cavern. Communication signals are blocked due to rocks in the way. This requires setting up communication relays. Hence, the robots need to cooperate in the form of a bucket brigade to establish a multi-hop communication link as shown in Figure 9 [4, 24]. The communication system has two fixed robots, one at the top of the cave (Base 0) and other at the base of the vertical entrance (Base 1). The Base 0 robot acts as 'base station' that receives data from all the robots inside the cave. The Base 1 robot acts as the intermediary that collects all the information from the other robots inside the cave and transmits it to the Base 0 robot. The remaining robots will perform exploration and at times organize into a bucket brigade establishing a multi-hop communication link from the farthest robot to the Base 0 robot. Each robot is equipped with a low power transmitter/receiver that has a range of 30 meters and uses 2.4 GHz S-band. Hence, a team of 36 SphereX robots with each robot acting as a relay can establish a communication link over a distance of 1 km.

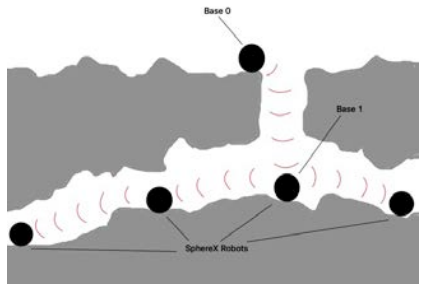

**Figure 9. Multi-hop communication link strategy**

**Mapping**

Each SphereX robot is equipped with 3 CMOS cameras, one on top, two attached to the turret. The top camera is used to map the top of the cave and the front camera on the turret can rotate $360^0$ at 1 rpm. The robot can take panoramic pictures and can scan the environment without having to use the propulsion system or reaction-wheels. Each camera has 1280×800 pixels and has a lens-field-of-view of $75^o$ horizontal and $47^o$ vertical. Based on the number of pixels and lens-field-of-view, we can calculate the range of the camera for a desired image resolution. The resolution of the camera at a depth $D$ in mm/pixel is defined as the observed area, $A$ divided by the number of pixels as shown below[23]:

$$R(D) = \frac{A(D)}{N_H N_V} = \frac{l_H l_V D^2}{N_H N_V f^2} \tag{6}$$

where, $N_H$ and $N_V$ are the number of pixels along the horizontal and vertical axes, $f$ is the focal length, $l_H$ and $l_V$ are the sensor height and length of the camera. Figure 10 shows the relation between resolution in mm/pixel and the depth for a camera of focal length 2.5 mm. As the distance between the camera and object increases, the resolution in mm/sec increases by the square law.

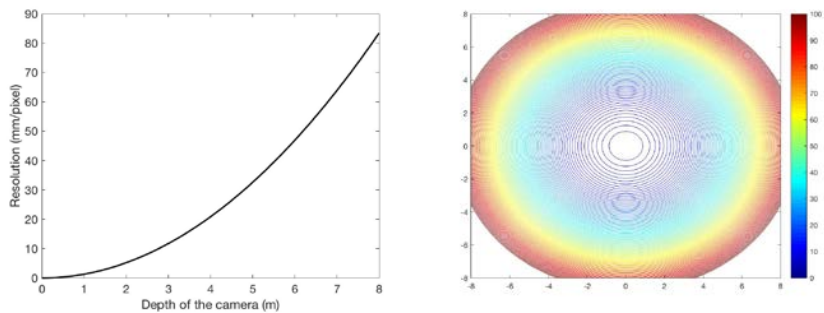

**Figure 10. Relation between resolution in mm/sec and depth of camera**

The minimum number of robots (*N*) required to map a length of a cave can be calculated by dividing the total length of the cave (*d*) by the maximum distance that two robots can be separated and adding the two base robots as shown below.

$$N = \frac{d}{30} + 2 \qquad (7)$$

The shortest time required to map a length of a cave at a desired resolution is a function of the total number of hops required to cover the area, time required for each hop and the time required to map the area by rotating the turret at 1 rpm as shown below.

$$T_h = \frac{d}{30} + \frac{d}{2R} \qquad (8)$$

$$T_{total} = \frac{d}{30}t_{h1} + \frac{d}{2R}(t_{h2}+1) + T_h \qquad (9)$$

where, $T_h$ is the total number of hops required to map the cave, $R$ is the range of the camera at the desired resolution, $t_{h1}$ is the time required to hop to cover the area at an increment of 30 m, $t_{h2}$ is the time required to hop to map the area at an increment of *2R*.

In Table 4, we analyze the number of robots and shortest time required to map a cave at a desired resolution. We consider a cave of 0.5 km, 1 km, 2 km and 5 km in length and the corresponding time required to map it within a resolution of 5 mm/pixel, 10 mm/pixel, 20 mm/pixel, 30 mm/pixel, 50 mm/pixel and 80 mm/pixel. We have considered the height and width of the cave to be 3 m and 4 m respectively. Note two successive robots can be separated by up to 30 m. The results show that for a reasonable size cave of 1 km or more requires dozens of robots to perform detailed mapping.

**Table 4. Number of robots and time required to map a cave**

| Resolution (mm/pixel) | Camera Range (m) | Cave Length | | | | | | | |
|---|---|---|---|---|---|---|---|---|---|
| | | 0.25 km | | 0.5 km | | 1 km | | 2 km | |
| | | Robots (#) | Map Time (min) | Robots (#) | Map Time (min) | Robots (#) | Map Time (min) | Robots (#) | Map Time (min) |
| 5 | 1.96 | 11 | 83 | 19 | 156 | 36 | 303 | 69 | 595 |
| 10 | 2.77 | 11 | 62 | 19 | 116 | 36 | 225 | 69 | 443 |
| 20 | 3.92 | 11 | 47 | 19 | 88 | 36 | 171 | 69 | 335 |
| 30 | 4.80 | 11 | 41 | 19 | 76 | 36 | 147 | 69 | 288 |
| 50 | 6.20 | 11 | 34 | 19 | 63 | 36 | 123 | 69 | 241 |
| 80 | 7.84 | 11 | 29 | 19 | 54 | 36 | 105 | 69 | 205 |

**DISCUSSION**

The proposed SphereX robots offer a compelling, practical solution that utilizes COTs technologies to provide access to extreme environments not possible with current planetary rovers. Despite significant research in the field, many conventional options are not practical for an off-world environment. Use of a miniature rocket system to propel the SphereX robot is simple, enabling hopping, short-flights and rolling. There are however developmental challenges in miniaturizing the rocket thrusters. Our analysis shows that several candidate propellants such as CDT(80) and RP1-Hydrogen Peroxide (diluted) offer reasonable solutions that trade performance with increased safety, storage and overall simplicity.

A team of SphereX robots maybe used to fully map a cave of a few kilometers in a few hours. This requires a team of dozens of robots to fan out, explore and then relay that data back to a base station. However, beyond a few kilometers, the number of robots required increases significantly into the hundreds. This suggests utilizing sensors such as LIDARs that can quickly scan and map topography, with limited coverage using visual and thermal cameras. However, in all scenarios it is critical that multiple robots are present both to avoid single point failures, in addition to relaying the date back to a base-station.

**CONCLUSION**

This paper presents the SphereX robot that uses rocket thrusters to hop, fly and roll in extreme off-world environments such as caves, lava-tubes and canyons. The proposed concept will allow mapping of these extreme environments using high resolution cameras. They offer the possibility of accessing these sites, never before possible and even performing sample return. We identify some of the shortest development pathways towards realizing this technology. Much of the SphereX platform will use COTS hardware. Further development is required in miniaturizing the selected propulsion system and towards coordination of robot teams. We also provided a brief summary of the dynamics of the SphereX system, navigation and path planning logistics. Our feasibility studies show that with sufficient resources, it is possible to advance the SphereX platform for a technical demonstration in a relevant environment with the future goal of incorporating the robots on a science-led surface mission to the Moon, Mars or asteroids.